\documentclass[10pt,twocolumn,letterpaper]{article}

\usepackage{cvpr}
\usepackage{times}
\usepackage{epsfig}
\usepackage{graphicx}
\usepackage{amsmath}
\usepackage{amssymb}
\usepackage{subcaption}
\usepackage{boldline,multirow}
\usepackage{comment}
\usepackage{algorithm}
\usepackage{algorithmicx}
\usepackage[noend]{algpseudocode}
\usepackage[pagebackref=true,breaklinks=true,letterpaper=true,colorlinks,bookmarks=false]{hyperref}
\usepackage{xr}
\cvprfinalcopy 

\DeclareMathOperator*{\argmax}{argmax}
\newcommand{\norm}[1]{\left\lVert#1\right\rVert}

\begin{document}

\title{Compositional Convolutional Neural Networks:\\
A Deep Architecture with Innate Robustness to Partial Occlusion}

\author{
	Adam Kortylewski
	\;\; Ju He
	\;\; Qing Liu
	\;\; Alan Yuille\\	
	Johns Hopkins University\\	
}

\maketitle

	\begin{abstract}
Recent findings show that deep convolutional neural networks (DCNNs) do not generalize well under partial occlusion.
Inspired by the success of compositional models at classifying partially occluded objects, we propose to integrate compositional models and DCNNs into a unified deep model with innate robustness to partial occlusion.
We term this architecture Compositional Convolutional Neural Network.
In particular, we propose to replace the fully connected classification head of a DCNN with a differentiable compositional model. 
The generative nature of the compositional model enables it to localize occluders and subsequently focus on the non-occluded parts of the object.
We conduct classification experiments on artificially occluded images as well as real images of partially occluded objects from the MS-COCO dataset.
The results show that DCNNs do not classify occluded objects robustly, even when trained with data that is strongly augmented with partial occlusions.
Our proposed model outperforms standard DCNNs by a large margin at classifying partially occluded objects, even when it has not been exposed to occluded objects during training. 
Additional experiments demonstrate that CompositionalNets can also localize the occluders accurately, despite being trained with class labels only.
The code used in this work is publicly available \footnote{\scriptsize{\url{https://github.com/AdamKortylewski/CompositionalNets}}}.
\end{abstract}
	\begin{figure}
    \centering
	\begin{subfigure}{.95\linewidth}
		\includegraphics[width=\linewidth]{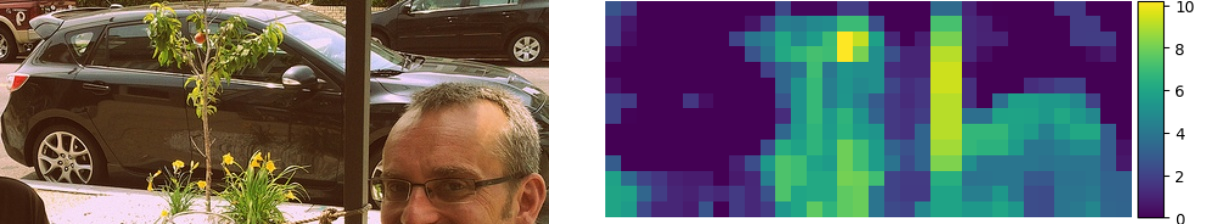}		
	\end{subfigure}
	\\\vspace{.1cm}
	\begin{subfigure}{.95\linewidth}
		\includegraphics[width=\linewidth]{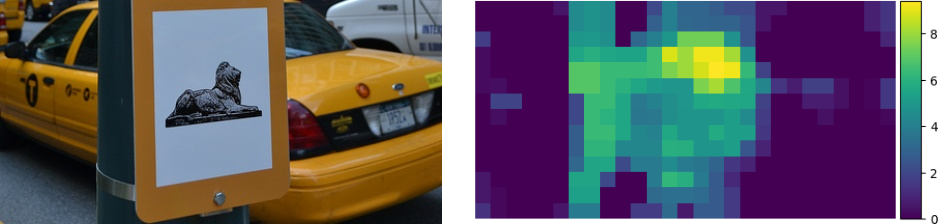}		
	\end{subfigure}	
	\\\vspace{.1cm}	
	\begin{subfigure}{.95\linewidth}
		\includegraphics[width=\linewidth]{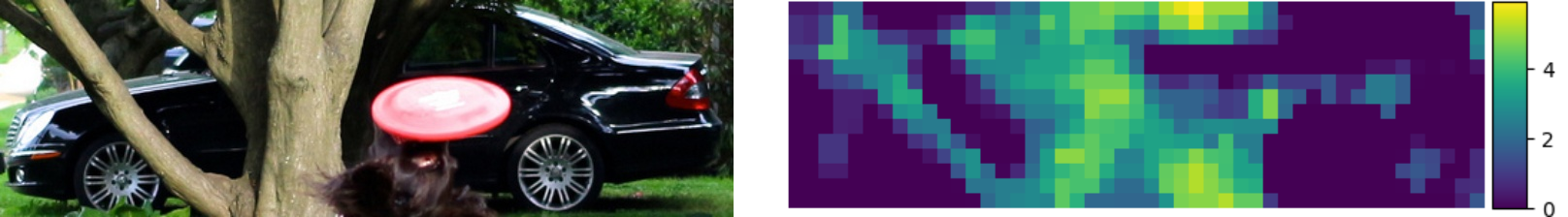}
	\end{subfigure}
	\caption{ 
		Partially occluded cars from the MS-COCO dataset \cite{lin2014microsoft} that are misclassified by a standard DCNN but correctly classified by the proposed CompositionalNet. Intuitively, a CompositionalNet can localize the occluders (occlusion scores on the right) and subsequently focus on the non-occluded parts of the object to classify the image.
		} 
	\label{fig:intro}
\end{figure}

\vspace{-0.15cm}
\section{Introduction}
Advances in the architecture design of deep convolutional neural networks (DCNNs) \cite{krizhevsky2012imagenet,simonyan2014very,he2016deep} 
increased the performance of computer vision systems at image classification enormously.
However, recent works \cite{hongru,kortylewski2019compositional} showed that such deep models are significantly less robust at classifying artificially occluded objects compared to Humans. 
Furthermore, our experiments show that DCNNs do not classify real images of partially occluded objects robustly. 
Thus, our findings and those of related works \cite{hongru,kortylewski2019compositional} point out a fundamental limitation of DCNNs in terms of generalization under partial occlusion which needs to be addressed.

One approach to overcome this limitation is to use data augmentation in terms of partial occlusion \cite{devries2017improved,yun2019cutmix}.
However, our experimental results show that after training with augmented data the performance of DCNNs at classifying partially occluded objects still remains substantially worse compared to the classification of non-occluded objects.

Compositionality is a  fundamental aspect of human cognition \cite{bienenstock1998compositionality,von1987synaptic,fodor1988connectionism,bienenstock1997compositionality} that is also reflected in the hierarchical compositional structure of the ventral stream in visual cortex \cite{yamane2008neural,vaziri2014channel,sasikumar2018first}.
A number of works in computer vision showed that compositional models can robustly classify partially occluded 2D patterns \cite{george2017generative,kortylewski2017model,wang2017detecting,zhang2018deepvoting}.
Kortylewski et al. \cite{kortylewski2019compositional} proposed dictionary-based compositional models, 
a generative model of neural feature activations that can classify images of partially occluded 3D objects more robustly than DCNNs.
However, their results also showed that their model is significantly less discriminative at classifying non-occluded objects compared to DCNNs. 

In this work, we propose to \textit{integrate} compositional models and DCNNs into a unified deep model with innate robustness to partial occlusion. 
In particular, we propose to replace the fully-connected classification head of a DCNN with a 
compositional layer that is regularized to be fully generative in terms of the neural feature activations of the last convolutional layer.
The generative property of the compositional layer enables the network to localize occluders in an image and subsequently focus on the non-occluded parts of the object in order to classify the image robustly.
We term this novel deep architecture Compositional Convolutional Neural Network (CompositionalNet).
Figure \ref{fig:intro} illustrates the robustness of CompositionalNets at classifying partially occluded objects, while also being able to localize occluders in an image.
In particular, it shows several images of cars that are occluded by other objects. 
Next to these images, we show occlusion scores that illustrate the position of occluders as estimated by the CompositionalNet.
Note how the occluders are accurately localized despite having highly complex shapes and appearances. 

Our extensive experiments demonstrate that the proposed CompositionalNet outperforms related approaches by a large margin at classifying partially occluded objects, even when it has not been exposed to occluded objects during training. 
When trained with data augmentation in terms of partial occlusion the performance increases further. 
In addition, we perform qualitative and quantitative experiments that demonstrate the ability of CompositionalNets to localize occluders accurately, despite being trained with class labels only.
We make several important contributions in this paper:
\begin{enumerate}
	\item We propose \textbf{a differentiable compositional model}  that is generative in terms of the feature activations of a DCNN .
	This enables us to integrate compositional models and deep networks into \textbf{compositional convolutional neural networks}, a unified deep model with innate robustness to partial occlusion.
	\item While previous works \cite{zhang2018deepvoting,kortylewski2019compositional,xiao2019tdapnet,hongru} evaluate robustness to partial occlusion on artificially occluded images only, we also \textbf{evaluate on real images of partially occluded objects} from the MS-COCO dataset.
	We demonstrate that CompositionalNets achieve \textbf{state-of-the-art results at classifying partially occluded objects} under occlusion.
	\item To the best of our knowledge we are the first to \textbf{study the task of localizing occluders} in an image and show that CompositionalNets outperform dictionary-based compositional models \cite{kortylewski2019compositional} substantially.
\end{enumerate}

		\section{Related Work}
	
	\textbf{Classification under partial occlusion.} 
	Recent work \cite{hongru,kortylewski2019compositional} has shown that current deep architectures are significantly less robust to partial occlusion compared to Humans.
	Fawzi and Frossard \cite{fawzi2016measuring} showed that DCNNs are vulnerable to partial occlusion simulated by masking small patches of the input image.
	Related works \cite{devries2017improved,yun2019cutmix}, have proposed to augment the training data with partial occlusion by masking out patches from the image during training. 
	However, our experimental results in Section \ref{sec:exp} show that such data augmentation approaches only have limited effects on the robustness of a DCNN to partial occlusion. 
	A possible explanation is the difficulty of simulating occlusion due to the large variability of occluders in terms of appearance and shape.
	Xiao et al. \cite{xiao2019tdapnet} proposed TDAPNet a deep network with an attention mechanism that masks out occluded features in lower layers to increase the robustness of the classification against occlusion.
	Our results show that this model does not perform well on images with real occlusion.
	In contrast to deep learning approaches, generative compositional models \cite{jin2006context,zhu2008,fidler2014,dai2014unsupervised,kortylewski2017greedy} have been shown to be inherently robust to partial occlusion when augmented with a robust occlusion model \cite{kortylewski2017model}.
	Such models have been successfully applied for detecting partially occluded object parts \cite{wang2017detecting,zhang2018deepvoting} and for recognizing 2D patterns under partial occlusion \cite{george2017generative,kortylewski2016probabilistic}.	
	
	\textbf{Combining compositional models and DCNNs.}  
	Liao et al. \cite{liao2016learning} proposed to integrate compositionality into DCNNs by regularizing the feature representations of DCNNs to cluster during learning. 
	Their qualitative results show that the resulting feature clusters resemble part-like detectors.
	Zhang et al. \cite{zhang2018interpretable} demonstrated that part detectors emerge in DCNNs by restricting the activations in feature maps to have a localized distribution. 
	However, these approaches have not been shown to enhance the robustness of deep models to partial occlusion. 
	Related works proposed to regularize the convolution kernels to be sparse \cite{tabernik2016towards}, or to force feature activations to be disentangled for different objects \cite{stone2017teaching}. 
 	As the compositional model is not explicit but rather implicitly encoded within the parameters of the DCNNs, the resulting models remain black-box DCNNs that are not robust to partial occlusion. 
 	A number of works \cite{li2019aognets,tang2018deeply,tang2017towards} use differentiable graphical models to integrate part-whole compositions into DCNNs. 
 	However, these models are purely discriminative and thus also are deep networks with no internal mechanism to account for partial occlusion.
	Kortylewski et al. \cite{kortylewski2019compositional} proposed learn a generative dictionary-based compositional models from the features of a DCNN.
	They use their compositional model as ``backup" to an independently trained DCNN, if the DCNNs classification score falls below a certain threshold.  
	
	In this work, we propose to integrate generative compositional models and DCNNs into a unified model that is inherently robust to partial occlusion.
    In particular, we propose to replace the fully connected classification head with a differentiable compositional model. 
    We train the model parameters with backpropagation, while regularizing the compositional model to be generative in terms of the neural feature activations of the last convolution layer.	
    Our proposed model significantly outperforms related approaches at classifying partially occluded objects while also being able to localize occluders accurately. 

	\section{Compositional Convolutional Neural Nets}
In Section \ref{sec:vmf}, we introduce a fully generative compositional model
and discuss how it can be integrated with DCNNs in an end-to-end system in Section \ref{sec:e2e}.

\subsection{Fully Generative Compositional Models}
\label{sec:vmf}
We denote a feature map $F^l\in \mathbb{R}^{H\times W \times D}$ as the output of a layer $l$ in a DCNN, with $D$ being the number of channels. A feature vector $f^l_p \in \mathbb{R}^D$ is the vector of features in $F^l$ at position $p$ on the 2D lattice $\mathcal{P}$ of the feature map.
In the remainder of this section we omit the superscript $l$ for notational clarity because this is fixed a-priori.

We propose a differentiable generative compositional model of the feature activations $p(F|y)$ for an object class $y$. 
This is different from dictionary-based compositional models \cite{kortylewski2019compositional} which learn a model $p(B|y)$, where $B$ is a non-differentiable binary approximation of $F$.
In contrast, we model the real-valued feature activations $p(F|y)$ as a mixture of von-Mises-Fisher (vMF) distributions:
\begin{align}
p(F|\theta_y) &=  \prod_{p} p(f_p|\mathcal{A}_{p,y},\Lambda) \label{eq:vmf}\\
p(f_p|\mathcal{A}_{p,y},\Lambda) &= \sum_k \alpha_{p,k,y} p(f_p|\lambda_k),\label{eq:vmf2}
\end{align}
where $\theta_y= \{\mathcal{A}_y,\Lambda\}$ are the model parameters and  $\mathcal{A}_y=\{\mathcal{A}_{p,y}\}$ are the parameters of the mixture models at every position $p \in \mathcal{P}$ on the 2D lattice of the feature map $F$. 
In particular, $\mathcal{A}_{p,y} = \{\alpha_{p,0,y},\dots,\alpha_{p,K,y}|\sum_{k=0}^K \alpha_{p,k,y} = 1\}$ are the mixture coefficients, $K$ is the number of mixture components and $\Lambda = \{\lambda_k = \{\sigma_k,\mu_k \} | k=1,\dots,K \}$ are the parameters of the vMF distribution:
\begin{equation}
\label{eq:vmfprob}
p(f_p|\lambda_k) = \frac{e^{\sigma_k \mu_k^T f_p}}{Z(\sigma_k)}, \norm{f_p} = 1, \norm{\mu_k}= 1, 
\end{equation}
where $Z(\sigma_k)$ is the normalization constant. 
The parameters of the vMF distribution $\Lambda$ can be learned by iterating between vMF clustering of the feature vectors of all training images and maximum likelihood parameter estimation \cite{banerjee2005clustering} until convergence. 
After training, the vMF cluster centers $\{\mu_k\}$ will resemble feature activation patterns that frequently occur in the training data. 
Interestingly, feature vectors that are similar to one of the vMF cluster centers, are often induced by image patches that are similar in appearance and often even share semantic meanings (see Supplementary A). 
This property was also observed in a number of related works that used clustering in the neural feature space \cite{wang2015unsupervised,liao2016learning,wang2017detecting}. 

The mixture coefficients $\alpha_{p,k,y}$ can also be learned with maximum likelihood estimation from the training images.
They describe the expected activation of a cluster center $\mu_k$ at a position $p$ in a feature map $F$ for a class $y$. 
Note that the spatial information from the image is preserved in the feature maps. 
Hence, our proposed vMF model (Equation \ref{eq:vmf}) intuitively describes the expected spatial activation pattern of parts in an image for a given class $y$ - e.g. where the tires of a car are expected to be located in an image.
In Section \ref{sec:e2e}, we discuss how the maximum likelihood estimation of the parameters $\theta_y$ can be integrated into a loss function and optimized with backpropagation.

\begin{figure*}
    \centering
    \includegraphics[width=\linewidth]{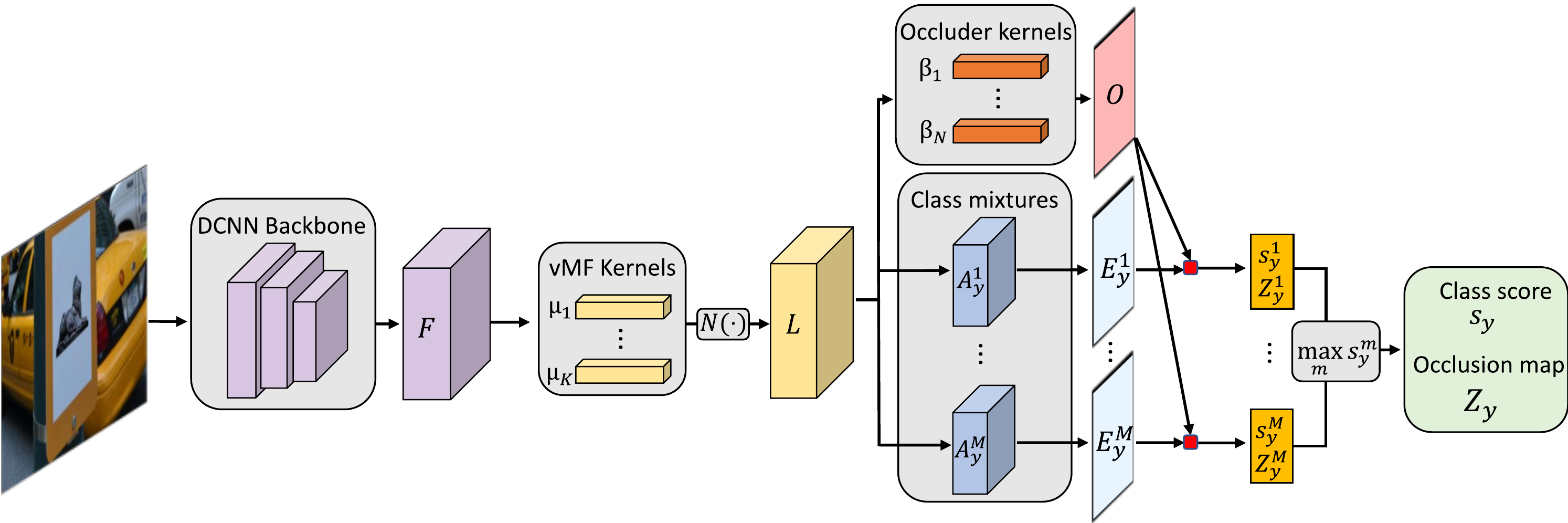}
    \caption{Feed-forward inference with a CompositionalNet. A DCNN backbone is used to extract the feature map $F$, followed by a convolution with the vMF kernels $\{\mu_k\}$ and a non-linear vMF activation function $\mathcal{N}(\cdot)$. The resulting vMF likelihood $L$ is used to compute the occlusion likelihood $O$ using the occluder kernels $\{\beta_n\}$.
    Furthermore, $L$ is used to compute the mixture likelihoods $\{E^m_y\}$ using the mixture models $\{A_y^{m}\}$. 
    $O$ and $\{E^m_y\}$ compete in explaining $L$ (red box) and are combined to compute an occlusion robust score $\{s^m_y\}$. The binary occlusion maps $\{Z^m_y\}$ indicate which positions in $L$ are occluded.
    The final class score $s_y$ is computed as $s_y\texttt{=}\max_m s^m_y$ and the occlusion map $Z_y$ is selected accordingly.}
    \label{fig:model}
\end{figure*}

\textbf{Mixture of compositional models.} 
The model in Equation \ref{eq:vmf} assumes that the 3D pose of an object is approximately constant in images. 
This is a common assumption of generative models that represent objects in image space.
We can represent 3D objects with a generalized model using mixtures of compositional models as proposed in \cite{kortylewski2019compositional}: 
\begin{equation}
\label{eq:mix}
p(F|\Theta_y) = \sum_m \nu^m p(F|\theta^m_y),
\end{equation}
with  $\mathcal{V}\texttt{=}\{\nu^m \in\{0,1\}, \sum_m \nu^m\texttt{=} 1 \}$ and $\Theta_y = \{\theta^m_y, m=1,\dots,M\}$. Here $M$ is the number of mixtures of compositional models and $\nu_m$ is a binary assignment variable that indicates which mixture component is active.
Intuitively, each mixture component $m$ will represent a different viewpoint of an object (see Supplementary B).
The parameters of the mixture components $\{\mathcal{A}^m_y\}$ need to be learned in an EM-type manner by iterating between estimating the assignment variables $\mathcal{V}$ and maximum likelihood estimation of $\{\mathcal{A}^m_y\}$. We discuss how this process can be performed in a neural network in Section \ref{sec:e2e}. 

\textbf{Occlusion modeling.}
Following the approach presented in \cite{kortylewski2017model}, compositional models can be augmented with an occlusion model. The intuition behind an occlusion model is that at each position $p$ in the image either the object model $p(f_p|\mathcal{A}^m_{p,y},\Lambda)$ or an occluder model $p(f_p|\beta,\Lambda)$ is active:
\begin{align}
	&p(F|\theta^m_y,\beta)\hspace{-0.075cm} =\hspace{-0.075cm} \prod_{p} p(f_p,z^m_p \hspace{0.05cm}\texttt{=}\hspace{0.05cm}0)^{1-z^m_p} p(f_p,z^m_p\hspace{0.05cm}\texttt{=}\hspace{0.05cm}1)^{z^m_p},\label{eq:occ}\\
	&p(f_p,z^m_p\hspace{0.05cm}\texttt{=}\hspace{0.05cm}1) = p(f_p|\beta,\Lambda)\hspace{0.1cm}p(z^m_p\texttt{=}1),\label{eq:occ2}\\
	&p(f_p,z^m_p\hspace{0.05cm}\texttt{=}\hspace{0.05cm}0) = p(f_p|\mathcal{A}^m_{p,y},\Lambda)\hspace{0.1cm}(1\texttt{-}p(z^m_p\texttt{=}1))\label{eq:occ3}.
\end{align}
The binary variables $\mathcal{Z}^m=\{z^m_p \in \{0,1\} | p \in \mathcal{P}\}$ indicate if the object is occluded at position $p$ for mixture component $m$. 
The occlusion prior $p(z^m_p\texttt{=}1)$ is fixed a-priori. 
Related works \cite{kortylewski2017model,kortylewski2019compositional} use a single occluder model. We instead use a mixture of several occluder models that are learned in an unsupervised manner: 
\begin{align}
p(f_p|\beta,\Lambda) &= \prod_n p(f_p|\beta_n,\Lambda)^{\tau_n} \\
&=\prod_n \Big(\sum_{k} \beta_{n,k} p(f_p|\sigma_k,\mu_k)\Big)^{\tau_n},
\end{align}
where \{$\tau_n \in \{0,1\},\sum_n \tau_n =1\}$ indicates which occluder model explains the data best.
The parameters of the occluder models $\beta_n$ are learned from clustered features of random natural images that do not contain any object of interest (see Supplementary C). 
Note that the model parameters $\beta$ are independent of the position $p$ in the feature map and thus the model has no spatial structure. 
Hence, the mixture coefficients $\beta_{n,k}$ intuitively describe the expected activation of $\mu_k$ anywhere in natural images. 

\textbf{Inference as feed-forward neural network.} 
The computational graph of our fully generative compositional model is directed and acyclic. 
Hence, we can perform inference in a single forward pass as illustrated in Figure \ref{fig:model}. 

We use a standard DCNN backbone to extract a feature representation $F = \psi(I,\omega) \in 
\mathbb{R}^{H \times W \times D}$ from the input image $I$, where $\omega$ are the parameters of the feature extractor.
The vMF likelihood function $p(f_p|\lambda_k)$ (Equation \ref{eq:vmfprob}) is composed of two operations: An inner product $i_{p,k}=\mu_k^T f_p$ and a non-linear transformation $\mathcal{N}=\exp(\sigma_k i_{p,k})/Z(\sigma_k)$.
Since $\mu_k$ is independent of the position $p$, computing $i_{p,k}$ is equivalent to a $1\times1$ convolution of $F$ with $\mu_k$. 
Hence, the vMF likelihood can be computed by:
\begin{equation}
L=\{\mathcal{N}(F\ast\mu_k)|k=1,\dots,K\} \in \mathbb{R}^{H\times W \times K}    
\end{equation}
(Figure \ref{fig:model} yellow tensor).
The mixture likelihoods $p(f_p|\mathcal{A}^m_{p,y},\Lambda)$ (Equation \ref{eq:vmf2}) are computed for every position $p$ as a dot-product between the mixture coefficients $\mathcal{A}^m_{p,y}$ and the corresponding vector $l_p\in\mathbb{R}^K$ from the likelihood tensor: 
\begin{equation}
E^m_y = \{l_p^T \mathcal{A}^m_{p,y} |\forall p\in\mathcal{P}\} \in \mathbb{R}^{H\times W},    
\end{equation}
(Figure \ref{fig:model} blue planes). 
Similarly, the occlusion likelihood can be computed as $O=\{\max_n l_p^T \beta_{n}| \forall p\in\mathcal{P}\}\in\mathbb{R}^{H \times W}$
(Figure \ref{fig:model} red plane).
Together, the occlusion likelihood $O$ and the mixture likelihoods $\{E^m_y\}$ are used to estimate the overall likelihood of the individual mixtures as $s^m_y=p(F|\theta_y^m,\beta)=\sum_p \max(E^m_{p,y},O_p)$. The final model likelihood is computed as $s_y= p(F|\Theta_y)=\max_m s^m_y$ and the final occlusion map is selected accordingly as $\mathcal{Z}_y = \mathcal{Z}^{\bar{m}}_y \in \mathbb{R}^{H \times W}$ where $\bar{m}=\argmax_m s^m_y$.

\subsection{End-to-end Training of CompositionalNets}
\label{sec:e2e}
We integrate our compositional model with DCNNs into \textit{Compositional Convolutional Neural Networks} (CompositionalNets) by replacing the classical fully connected classification head with a compositional model head as illustrated in Figure \ref{fig:model}. 
The model is fully differentiable and can be trained end-to-end using backpropagation. 
Algorithm \ref{alg:train} shows the initialization and training of our CompositionalNets as pseudo code.
The trainable parameters of a CompositionalNet are $T=\{\omega,\Lambda,\mathcal{A}_y\}$.
We optimize those parameters jointly using stochastic gradient descent. The loss function is composed of four terms:
\begin{align}
    \mathcal{L}(y,y',F,T) =  & \mathcal{L}_{class}(y,y') + \gamma_1 \mathcal{L}_{weight}(\omega) +\\
                 & \gamma_2 \mathcal{L}_{vmf}(F,\Lambda) + \gamma_3 \mathcal{L}_{mix}(F,\mathcal{A}_y).
\end{align}
$\mathcal{L}_{class}(y,y')$ is the cross-entropy loss between the network output $y'$ and the true class label $y$. $\mathcal{L}_{weight} = \norm{\omega}^2_2$ is a weight regularization on the DCNN parameters.
$\mathcal{L}_{vmf}$ and $\mathcal{L}_{mix}$ regularize the parameters of the compositional model to have maximal likelihood for the features in $F$.
$\{\gamma_1,\gamma_2,\gamma_3\}$ control the trade-off between the loss terms. 

\begin{algorithm}[t]
    {\small
	\textbf{Input:} Set of training images $I=\{I_1,\dots,I_H\}$, \\ 
	labels $y=\{y_1,\dots,y_H\}$, VGG backbone $\psi(\cdot,\omega)$,\\
	background images $B=\{B_1,\dots,B_R\}$.\\
	\textbf{Output:} Model parameters $T=\{\omega,\{\mu_k\},\{\mathcal{A}^m_y\}\}$,$\{\beta_n\}$.
	\begin{algorithmic}[1]
	    \State \texttt{//extract features}
		\State $\{F_h\}$ $\gets$ $\psi(\{I_h\},\omega)$ 
		\State \texttt{//initialize vMF kernels by ML}
		\State $\{\mu_k\}$$\gets$cluster\_and\_ML($\{f_{h,p}|h$=$\{1,$...$,H\},p\hspace{-.05cm}\in\hspace{-.05cm}\mathcal{P}\}$)
		\State $\{L_h\}$ $\gets$ compute\_vMF\_likelihood($\{F_h\},\{\mu_k\}$) [Eq. 10]
		\State \texttt{//initialize mixture models by ML}
		\State $\{\mathcal{A}^m_y\}$ $\gets$ cluster\_and\_ML($\{L_h\},y$)
		\State $\{\beta_n\}$ $\gets$ learn\_background\_models($B$,$\psi(\cdot,\omega)$,$\{\mu_k\}$)
		\For{\#epochs}
		    \For{each image $I_h$}
		        \State \hspace{-.2cm} $\{y'_h,m^\uparrow,\{z_p^\uparrow\}\}$ $\gets$ inference($I_h,T,\{\beta_n\}$) 
		        \State \hspace{-.2cm} $T$ $\gets$ optimize($y_h$,$y'_h$,$\omega$,$\{\mu_k\}$,$\mathcal{A}^{m^\uparrow}_{y}$,$\{z_p^\uparrow\}$) [Sec. 3.2]
		        \EndFor
		    \EndFor
	\end{algorithmic}	
	}
	\caption{Training of CompositionalNets}\label{alg:train}	
\end{algorithm}

The vMF cluster centers $\mu_k$ are learned by maximizing the vMF-likelihoods (Equation \ref{eq:vmfprob}) for the feature vectors $f_p$ in the training images. We keep the vMF variance $\sigma_k$ constant, which also reduces the normalization term $Z(\sigma_k)$ to a constant. 
We assume a hard assignment of the feature vectors $f_p$ to the vMF clusters during training. Hence, the free energy to be minimized for maximizing the vMF likelihood \cite{wang2017visual} is:
\begin{align}
    \mathcal{L}_{vmf}(F,\Lambda)  &= - \sum_p \max_k \log p(f_p|\mu_k)\\
                        &= C \sum_p \min_k \mu_k^T f_p,
\end{align}
where $C$ is a constant. Intuitively, this loss encourages the cluster centers $\mu_k$ to be similar to the feature vectors $f_p$. 

In order to learn the mixture coefficients $\mathcal{A}^m_{y}$ we need to maximize the model likelihood (Equation \ref{eq:mix}). 
We can avoid an iterative EM-type learning procedure by making use of the fact that the the mixture assignment $\nu_m$ and the occlusion variables $z_{p}$ have been inferred in the forward inference process.
Furthermore, the parameters of the occluder model are learned a-priori and then fixed. Hence the energy to be minimized for learning the mixture coefficients is:
\begin{equation}
    \mathcal{L}_{mix}(F,\mathcal{A}_y) =\hspace{-.05cm}\texttt{-}\hspace{-.1cm}\sum_p\hspace{-0.05cm}(1\texttt{-}z^\uparrow_{p}) \log \hspace{-0.05cm}\Big[\hspace{-0.05cm}\sum_k\hspace{-0.1cm}\alpha^{m^\uparrow}_{p,k,y}p(f_p|\lambda_k)\Big]
\end{equation}
Here, $z^\uparrow_{p}$ and $m^\uparrow$ denote the variables that were inferred in the forward process (Figure \ref{fig:model}). 

	\begin{figure}
	\begin{subfigure}{0.3\linewidth}
		\centering
		\includegraphics[height=1.25cm]{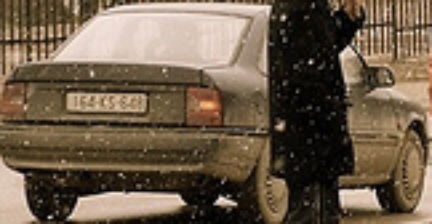}
	\end{subfigure}%
	\begin{subfigure}{0.39\linewidth}
		\centering
		\includegraphics[height=1.25cm]{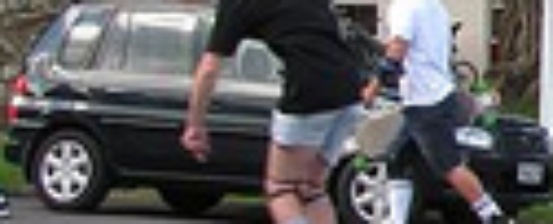}
	\end{subfigure}%
	\begin{subfigure}{0.3\linewidth}
		\centering
		\includegraphics[height=1.25cm]{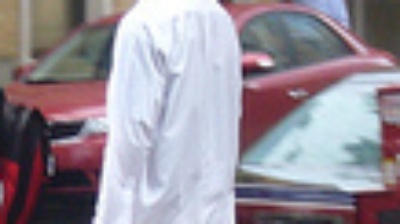}
	\end{subfigure}%
	\vspace{.1cm}
	\\\\\
	\vspace{.1cm}
	\begin{subfigure}{0.33\linewidth}
		\centering
		\includegraphics[height=1.55cm]{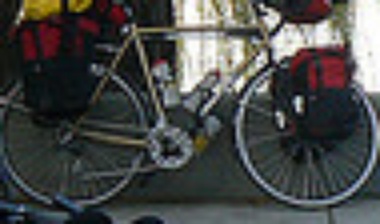}
	\end{subfigure}%
	\begin{subfigure}{0.33\linewidth}
		\centering
		\includegraphics[height=1.55cm]{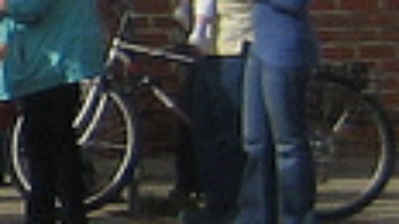}
	\end{subfigure}%
	\begin{subfigure}{0.33\linewidth}
		\centering
		\includegraphics[height=1.55cm]{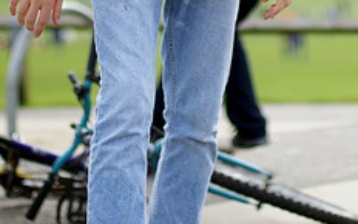}
	\end{subfigure}%
	\\\\\
	\begin{subfigure}{0.37\linewidth}
		\centering
		\includegraphics[height=1.65cm]{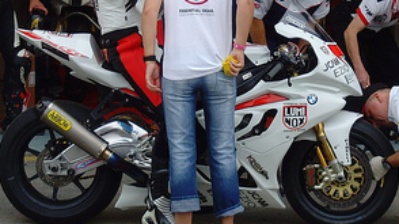}
	\end{subfigure}%
	\begin{subfigure}{0.3\linewidth}
		\centering
		\includegraphics[height=1.65cm]{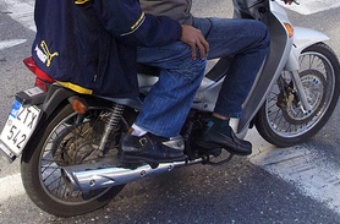}
	\end{subfigure}%
	\begin{subfigure}{0.32\linewidth}
		\centering
		\includegraphics[height=1.65cm]{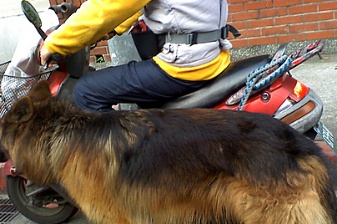}
	\end{subfigure}%
    \caption{Images from the Occluded-COCO-Vehicles dataset. Each row shows samples of one object class with increasing amount of partial occlusion: $20$-$40\%$ (Level-1), $40$-$60\%$ (Level-2), $60$-$80\%$ (Level-3).}
    \label{fig:occ-real}
\end{figure}

\begin{table*}
	\small
	\centering
	\tabcolsep=0.11cm
	\begin{tabular}{V{2.5}lV{2.5}cV{2.5}c|c|c|cV{2.5}c|c|c|cV{2.5}c|c|c|cV{2.5}cV{2.5}}
		\multicolumn{15}{c}{\textbf{PASCAL3D+ Vehicles Classification under Occlusion}} \\
		\hline
		Occ. Area 					         & \textbf{L0: 0\%} & \multicolumn{4}{cV{2.5}}{\textbf{L1: 20-40\%}} & \multicolumn{4}{cV{2.5}}{\textbf{L2: 40-60\%}} & \multicolumn{4}{cV{2.5}}{\textbf{L3: 60-80\%}} & Mean \\
		\hline
		Occ. Type 	& - & w & n & t & o & w & n & t& o & w & n & t & o & \\    		
		\hline  
		VGG 
		& 99.2&96.9&97.0&96.5&93.8&92.0&90.3&89.9&79.6&67.9&62.1&59.5&62.2&83.6\\
		CoD\cite{kortylewski2019compositional}	&92.1&92.7&92.3&91.7&92.3&87.4&89.5&88.7&90.6&70.2&80.3&76.9&87.1&87.1\\
		VGG+CoD \cite{kortylewski2019compositional}
		&98.3&96.8&95.9&96.2&94.4&91.2&91.8&91.3&91.4&71.6&80.7&77.3&87.2&89.5\\
		TDAPNet \cite{xiao2019tdapnet}
		&\textbf{99.3}&98.4&98.6&98.5&97.4&96.1&97.5&96.6&91.6&82.1&88.1&82.7&79.8&92.8\\		
		\hline	
		CompNet-p4&97.4&96.7&96.0&95.9&95.5&95.8&94.3&93.8&92.5&86.3&84.4&82.1&88.1&92.2\\
		CompNet-p5&\textbf{99.3}&98.4&\textbf{98.6}&98.4&96.9&98.2&98.3&97.3&88.1&90.1&89.1&83.0&72.8&93.0\\
		CompNet-Multi&\textbf{99.3}&\textbf{98.6}&\textbf{98.6}&\textbf{98.8}&\textbf{97.9}&\textbf{98.4}&\textbf{98.4}&\textbf{97.8}&\textbf{94.6}&\textbf{91.7}&\textbf{90.7}&\textbf{86.7}&\textbf{88.4}&\textbf{95.4}\\
	    \hline
	\end{tabular}
	\caption{Classification results for vehicles of PASCAL3D+ with different levels of artificial occlusion (0\%,20-40\%,40-60\%,60-80\% of the object are occluded) and different types of occlusion (w=white boxes, n=noise boxes, t=textured boxes, o=natural objects). CompositionalNets outperform related approaches significantly.}
	\label{tab:p3d}
\end{table*}

\begin{table*}
	\small
	\centering
	\tabcolsep=0.11cm
	\begin{tabular}{V{2.5}lV{2.5}c|c|c|c|cV{2.5}c|c|c|c|cV{2.5}c|c|c|c|cV{2.5}c|c|c|c|cV{2.5}}
		\multicolumn{21}{c}{\textbf{MS-COCO Vehicles Classification under Occlusion}} \\
		\hline Train Data & \multicolumn{5}{cV{2.5}}{\textbf{PASCAL3D+}}
		                      & \multicolumn{5}{cV{2.5}}{\textbf{MS-COCO}}
		                      & \multicolumn{5}{cV{2.5}}{\textbf{MS-COCO + CutOut}}
		                      & \multicolumn{5}{cV{2.5}}{\textbf{MS-COCO + CutPaste}}\\
		\hline Occ. Area & \textbf{L0} & \textbf{L1} & \textbf{L2}& \textbf{L3}& Avg
		                 & \textbf{L0} & \textbf{L1} & \textbf{L2}& \textbf{L3}& Avg
		                 & \textbf{L0} & \textbf{L1} & \textbf{L2}& \textbf{L3}& Avg
		                 & \textbf{L0} & \textbf{L1} & \textbf{L2}& \textbf{L3}& Avg\\
		\hline
		VGG &97.8&86.8&79.1&60.3&81.0
		    &99.1&88.7&78.8&63.0&82.4
		    &99.3&90.9&87.5&75.3&88.3
		    &99.3&92.3&89.9&80.8&90.6\\
		\hline
		CoD &91.8&82.7&83.3&76.7&83.6&-&-&-&-&-&-&-&-&-&-&-&-&-&-&-\\
		\hline
		VGG+CoD&98.0&88.7&80.7&69.9&84.3&-&-&-&-&-&-&-&-&-&-&-&-&-&-&-\\
		\hline
		TDAPNet&98.0&88.5&85.0&74.0&86.4&\textbf{99.4}& 88.8& 87.9& 69.9&86.5&99.3 & 90.1 & 88.9 & 71.2&87.4&98.1 & 89.2 & 90.5 & 79.5&89.3\\
		\hline
		CompNet-p4&96.6&91.8&85.6&76.7&87.7&97.7&92.2&86.6&82.2&89.7&97.8&91.9&87.6&79.5&89.2&98.3&93.8&88.6&84.9&91.4\\
		\hline
		CompNet-p5&98.2&89.1&84.3&78.1&87.5&99.1&92.5&87.3&82.2&90.3&99.3&93.2&87.6&84.9&91.3&\textbf{99.4}&93.9&90.6&\textbf{90.4}&93.5\\
		\hline
		CompNet-Mul&\textbf{98.5}&\textbf{93.8}&\textbf{87.6}&\textbf{79.5}&\textbf{89.9}&\textbf{99.4}&\textbf{95.3}&\textbf{90.9}&\textbf{86.3}&\textbf{93.0}&\textbf{99.4}&\textbf{95.2}&\textbf{90.5}&\textbf{86.3}&\textbf{92.9}&\textbf{99.4}&\textbf{95.8}&\textbf{91.8}&\textbf{90.4}&\textbf{94.4}\\
		\hline
	\end{tabular}
	\caption{Classification results for vehicles of MS-COCO with different levels of real occlusion (L0: 0\%,L1: 20-40\%,L2 40-60\%, L3:60-80\% of the object are occluded). 
	The training data consists of images from: PASCAL3D+, MS-COCO as well as data from MS-COCO that was augmented with CutOut and CutPaste.
	CompositionalNets outperform related approaches in all test cases.}
	\label{tab:coco}	
\end{table*}

\section{Experiments}
\label{sec:exp}
We perform experiments at the tasks of classifying partially occluded objects and at occluder localization.

\begin{figure*}
	\begin{subfigure}{0.49\linewidth}
		\centering
		\includegraphics[height =1.3cm]{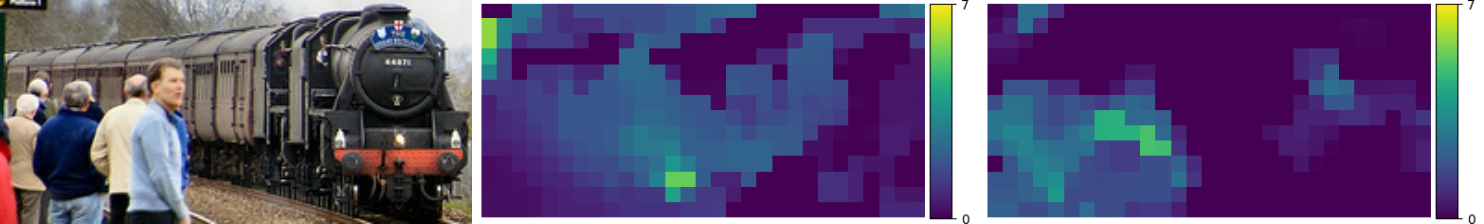}
	\end{subfigure}
	\begin{subfigure}{0.49\linewidth}
		\centering
		\includegraphics[height =1.2cm]{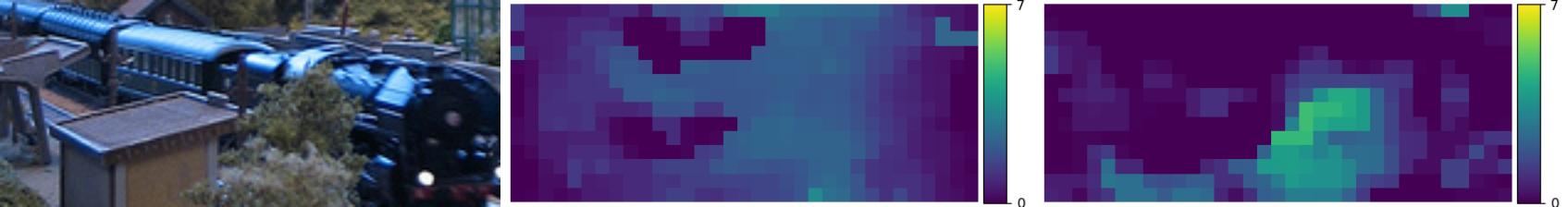}
	\end{subfigure}
	\begin{subfigure}{0.49\linewidth}
		\centering
		\includegraphics[width=\linewidth]{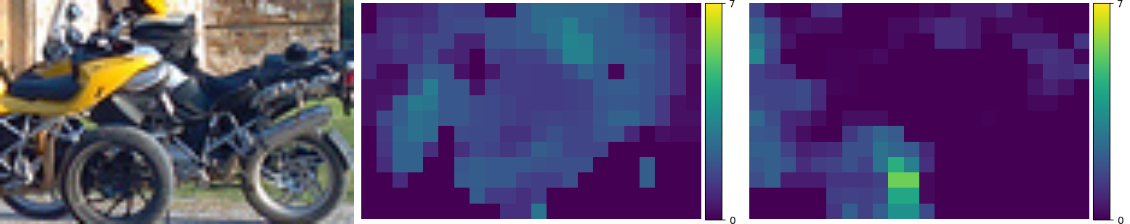}
	\end{subfigure}
	\begin{subfigure}{0.49\linewidth}
		\centering
		\includegraphics[width=\linewidth]{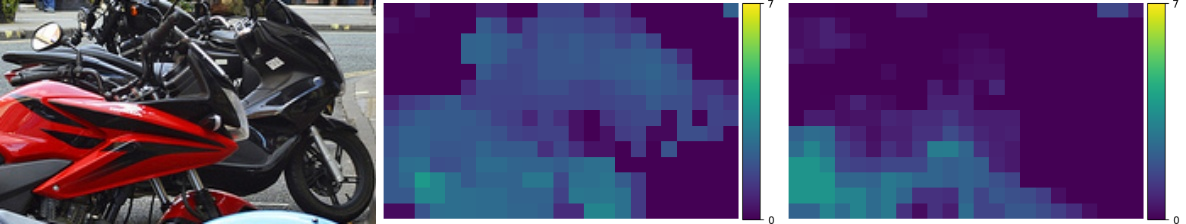}
	\end{subfigure}
	\begin{subfigure}{0.49\linewidth}
		\centering
		\includegraphics[width=\linewidth]{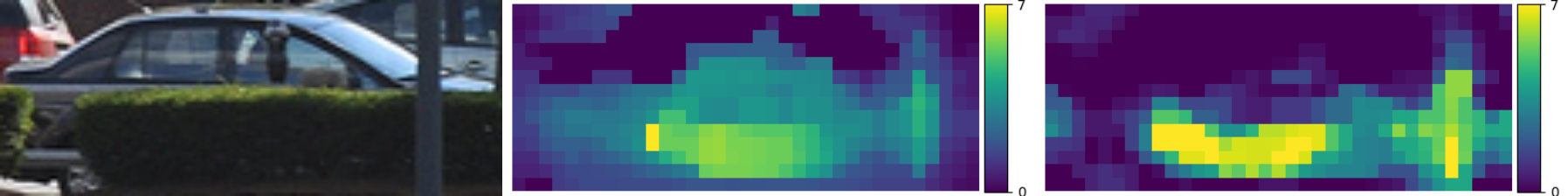}
	\end{subfigure}
	\begin{subfigure}{0.49\linewidth}
		\centering
		\includegraphics[width=\linewidth]{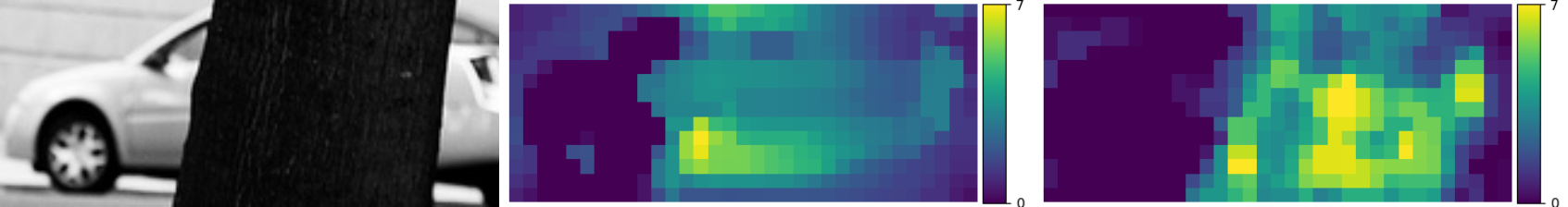}
	\end{subfigure}
	\begin{subfigure}{0.49\linewidth}
		\centering
		\includegraphics[width=\linewidth]{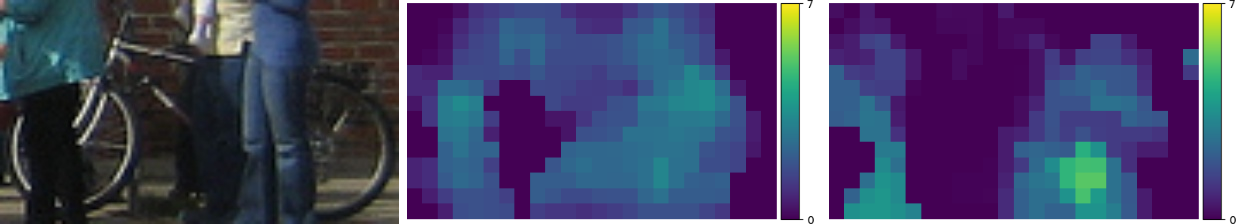}
	\end{subfigure}
	\begin{subfigure}{0.49\linewidth}
		\centering
		\includegraphics[width=\linewidth]{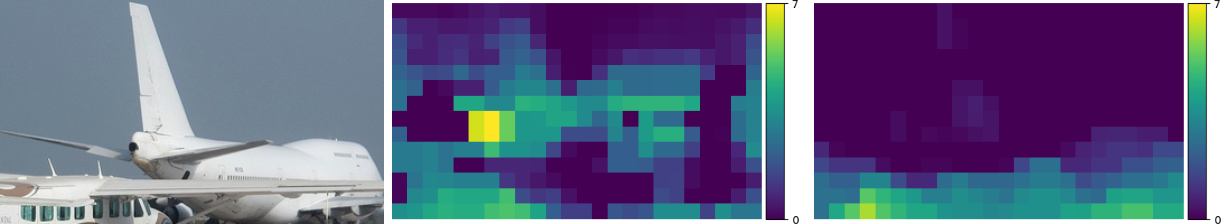}
	\end{subfigure}		
	\begin{subfigure}{0.49\linewidth}
		\centering
		\includegraphics[width=\linewidth]{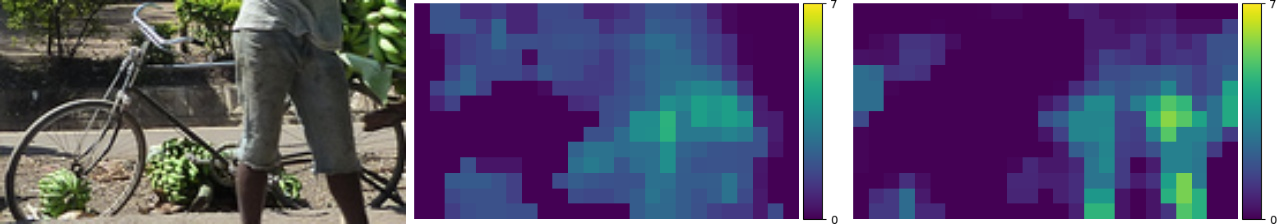}
	\end{subfigure}	
	\begin{subfigure}{0.49\linewidth}
		\centering
		\includegraphics[width=\linewidth]{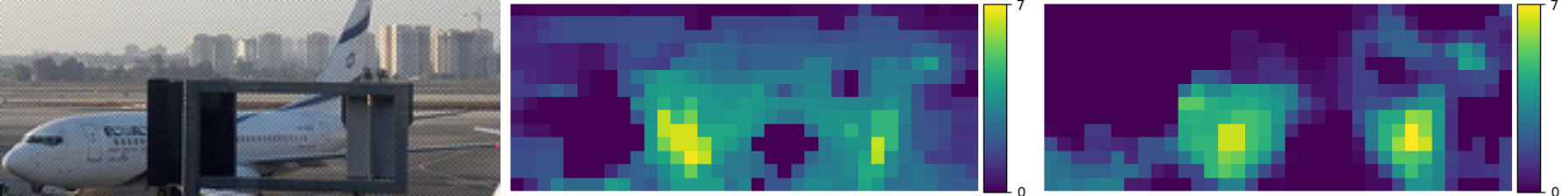}
	\end{subfigure}	
	\begin{subfigure}{0.49\linewidth}
		\centering
		\includegraphics[height=1.9cm]{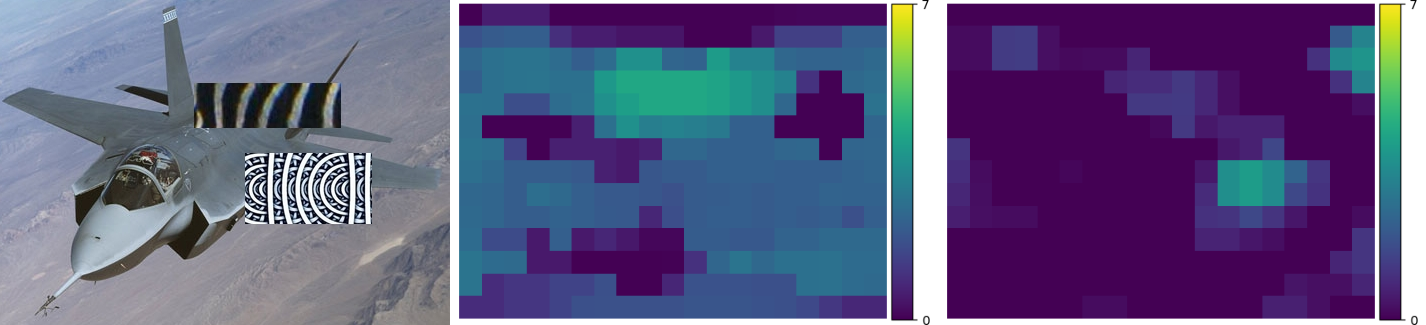}
	\end{subfigure}					
	\begin{subfigure}{0.49\linewidth}
		\centering
		\includegraphics[height =1.9cm]{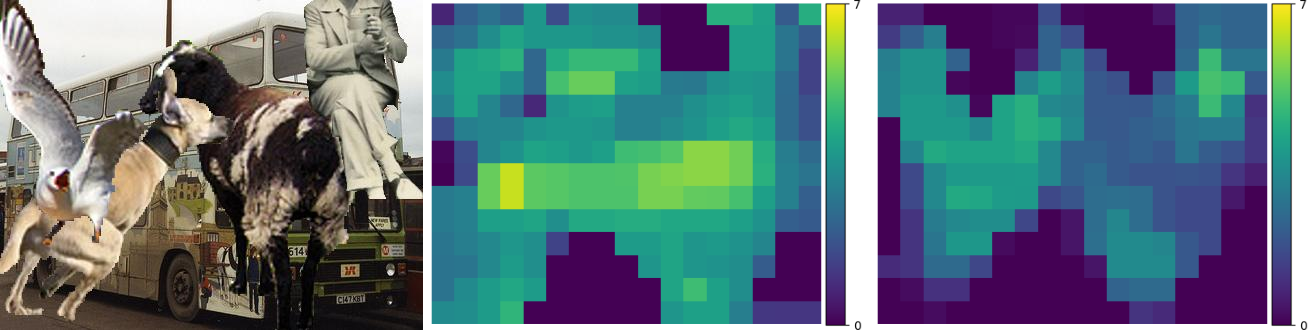}
	\end{subfigure}
	\begin{subfigure}{0.49\linewidth}
		\centering
		\includegraphics[height=2.2cm]{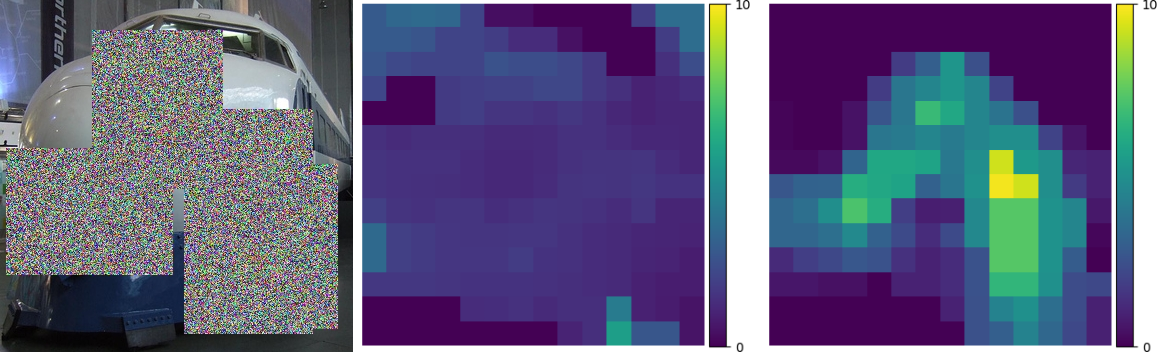}
	\end{subfigure}		
	\begin{subfigure}{0.49\linewidth}
		\centering
		\includegraphics[height=2.2cm]{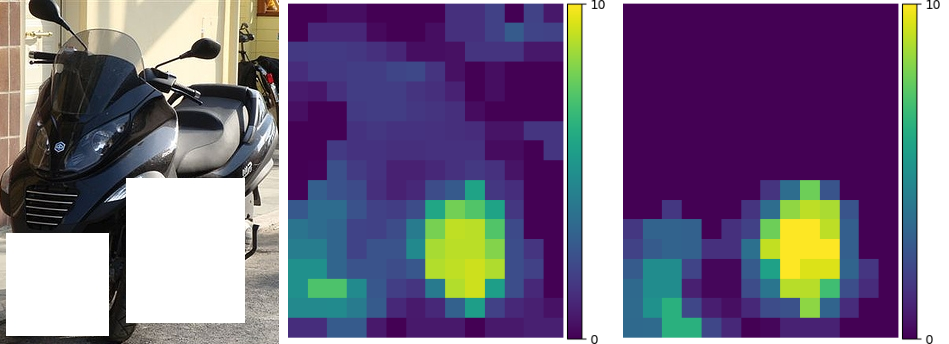}
	\end{subfigure}						
		
	\caption{Qualitative occlusion localization results. Each result consists of three images: The input image, and the occlusion scores of a dictionary-based compositional model \cite{kortylewski2019compositional} and our proposed CompositionalNet. Note how our model can localize occluders with higher accuracy across different objects and occluder types for real as well as for artificial occlusion.}
	\label{fig:occ-quali}
\end{figure*}

\textbf{Datasets.} For evaluation we use the \textit{Occluded-Vehicles} dataset as proposed in \cite{wang2015unsupervised} and extended in \cite{kortylewski2019compositional}. The dataset consists of images and corresponding segmentations of vehicles from the PASCAL3D+ dataset \cite{xiang2014beyond} that were synthetically occluded with four different types of occluders: segmented \textit{objects} as well as patches with \textit{constant white color}, \textit{random noise} and \textit{textures} (see Figure \ref{fig:occ-quali} for examples). 
The amount of partial occlusion of the object varies in four different levels: $0\%$ (L0), $20$-$40\%$ (L1), $40$-$60\%$ (L2), $60$-$80\%$ (L3).

While it is reasonable to evaluate occlusion robustness by testing on artificially generated occlusions, it is necessary to study the performance of algorithms under realistic occlusion as well. 
Therefore, we introduce a dataset with images of real occlusions which we term \textit{Occluded-COCO-Vehicles}. It consists of the same classes as the Occluded-Vehicle dataset. The images were generated by cropping out objects from the MS-COCO \cite{lin2014microsoft} dataset based on their bounding box. 
The objects are categorized into the four occlusion levels defined by the Occluded-Vehicles dataset based on the amount of the object that is visible in the image (using the segmentation masks available in both datasets).
The number of test images per occlusion level are: $2036$ (L0), $768$ (L1), $306$ (L2), $73$ (L3). For training purpose, we define a separate training dataset of $2036$ images from level L0. 
Figure \ref{fig:occ-real} illustrates some example images from this dataset.

\textbf{Training setup.} CompositionalNets are trained from the feature activations of a VGG-16 \cite{simonyan2014very} model that is pretrained on ImageNet\cite{deng2009imagenet}. We initialize the compositional model parameters $\{\mu_k\},\{\mathcal{A}_y\}$ using clustering as described in Section \ref{sec:vmf} and set the vMF variance to $\sigma_k = 30, \forall k \in \{1,\dots,K\}$. We train the model parameters $\{\{\mu_k\},\{\mathcal{A}_y\}\}$ using backpropagation. 
We learn the parameters of $n=5$ occluder models $\{\beta_1,\dots,\beta_n\}$ in an unsupervised manner as described in Section \ref{sec:vmf} and keep them fixed throughout the experiments. We set the number of mixture components to $M=4$. The mixing weights of the loss are chosen to be: $\gamma_1=0.1$, $\gamma_2=5$, $\gamma_3=1$. We train for $60$ epochs using stochastic gradient descent with momentum $r=0.9$ and a learning rate of $lr=0.01$. 

\subsection{Classification under Partial Occlusion}
\label{sec:exp-class}

\textbf{PASCAL3D+.} In Table \ref{tab:p3d} we compare our CompositionalNets to a VGG-16 network that was pre-trained on ImageNet and fine-tuned with the respective training data. Furthermore, we compare to a dictionary-based compositional model (CoD) and a combination of both models (VGG+CoD) as reported in \cite{kortylewski2019compositional}. 
We also list the results of TDAPNet as reported in \cite{xiao2019tdapnet}.
We report results of CompositionalNets learned from the \texttt{pool4} and \texttt{pool5} layer of the VGG-16 network respectively (CompNet-p4 \& CompNet-p5), as well as as a multi-layer CompositionalNet (CompNet-Multi) that is trained by combining the output of CompNet-p4 and CompNet-p5. 
In this setup, all models are trained with non-occluded images ($L0$), while at test time the models are exposed to images with different amount of partial occlusion ($L0$-$L3$). 

We observe that CompNet-p4 and CompNet-p5 outperform VGG-16, CoD as well as the combination of both significantly.
Note how the CompositionalNets are much more discriminative at level $L0$ compared to dictionary-based compositional models.
While CompNet-p4 and CompNet-p5 perform on par with the TDAPNet, CompNet-Multi outperforms TDAPNet significantly. 
We also observe that CompNet-p5 outperforms CompNet-p4 for low occlusions ($L0$ \& $L1$) and for stronger occlusions if the occluders are rectangular masks. 
However, CompNet-p4 outperforms CompNet-p5 at strong occlusion ($L2$ \& $L3$) when the occluders are objects. 
As argued by Xiao et al. \cite{xiao2019tdapnet} this could be attributed to the fact that occluders with more fine-grained shapes disturb the features in higher layers more severely.

\textbf{MS-COCO.}  Table \ref{tab:coco} shows classification results under a realistic occlusion scenario by testing on the Occluded-COCO-Vehicles dataset. 
The models in the first part of the Table are trained on non-occluded images of the PASCAL3D+ data and evaluated on the MS-COCO data.
While the performance drops for all models in this transfer learning setting, CompositionalNets outperform the other approaches significantly. 
Note that combining a DCNN with the dictionary-based compositional model (VGG+CoD) performs well at low occlusion $L0\&L1$ but lower performance for $L2\&L3$ compared to CoD only. 

The second part of the table (MS-COCO) shows the classification performance after fine-tuning on the $L0$ training set of the Occluded-COCO-Vehicles dataset. 
VGG-16 achieves a similar performance as for the artificial object occluders in Table \ref{tab:p3d}. 
After fine-tuning, TDAPNet improves at level $L0$ and decreases on average for the levels $L1-3$. 
Overall it does not significantly benefit from fine-tuning with non-occluded images.
The performance of the CompositionalNet increases substantially (p4: $3\%$, p5: $2.8\%$, multi: $3.1\%$) after fine-tuning. 

The third and fourth parts of Table \ref{tab:coco} (MS-COCO-CutOut \& MS-COCO-CutPaste) show classification results after training with strong data augmentation in terms of partial occlusion.
In particular, we use CutOut \cite{devries2017improved} regularization by masking out random square patches of size $70$ pixels. Furthermore, we propose a stronger data augmentation method \textit{CutPaste} which artificially occludes the training images in the Occluded-COCO-Vehicles dataset with all four types of artificial occluders used in the OccludedVehicles dataset.
While data augmentation increases the performance of the VGG network, the model still suffers from strong occlusions and falls below the CompNet-Multi model that was only trained on non-occluded images.
TDAPNet does not benefit from data augmentation as much as the VGG network. 
For CompositionalNets the performance increases further when trained with augmented data. 
Overall, CutOut augmentation does not have a large effect on the generalization performance of CompositionalNets, while the proposed CutPaste augmentation proves to be stronger.
In particular, the CompNet-p5 architecture benefits strongly, possibly because the network learns to extract more reliable higher level features under occlusion.

In summary, the classification experiments clearly highlight the robustness of CompositionalNets at classifying partially occluded objects, while also being highly discriminative when objects are not occluded. 
Overall CompositionalNets significantly outperform dictionary-based compositional models and other neural network architectures at image classification under artificial as well as real occlusion in all three tested setups - at transfer learning between datasets, when trained with non-occluded images and when trained with strongly augmented data. 

\begin{figure}
	\centering
	\includegraphics[width=0.9\linewidth]{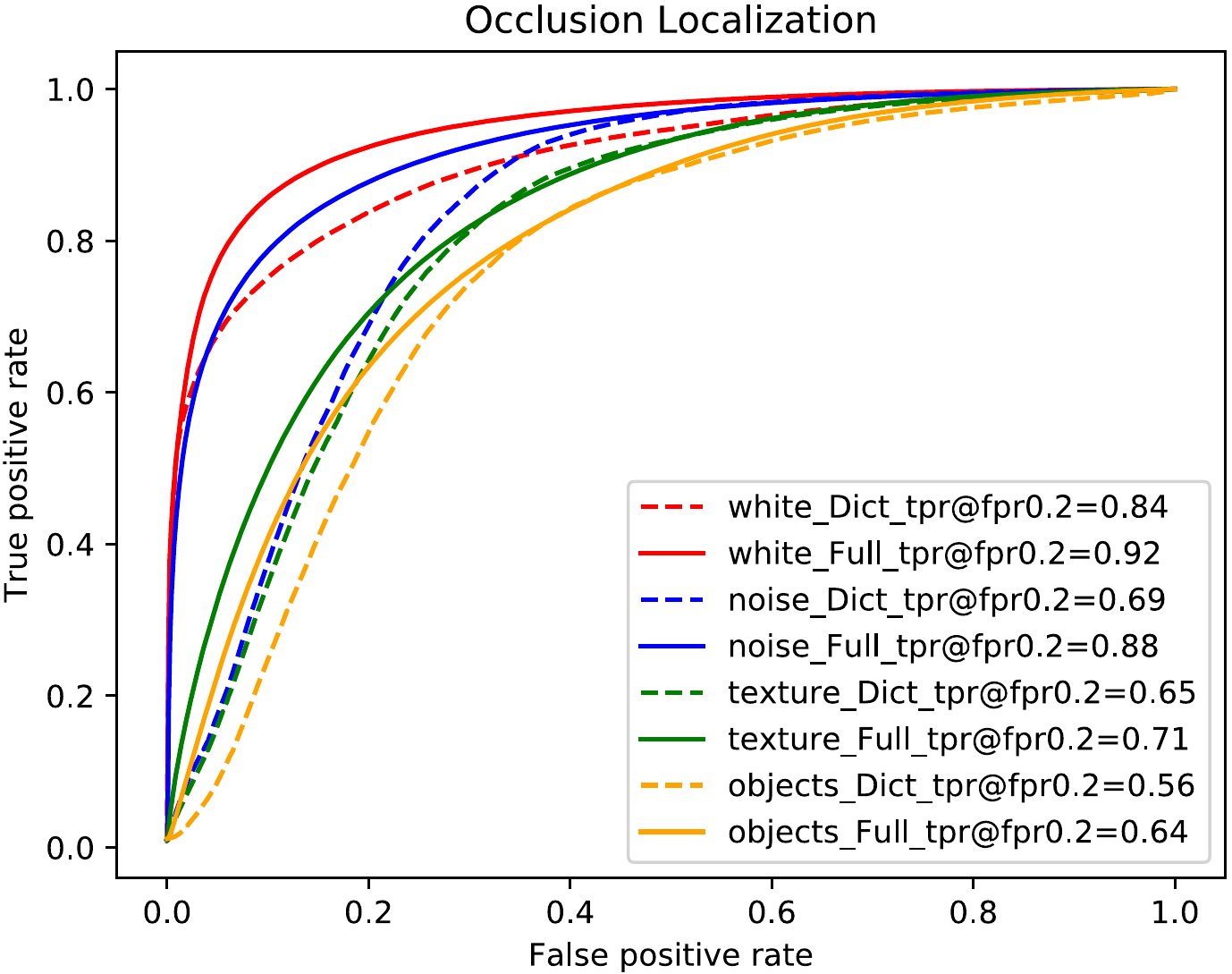}
	\caption{ROC curves for occlusion localization with dictionary-based compositional models and the proposed CompositionalNets averaged over all levels of partial occlusion (L1-L3). CompositionalNets significantly outperform dictionary-based compositional models.}
	\label{fig:occ-det}
	\vspace{-.25cm}
\end{figure}

\subsection{Occlusion Localization}
While it is important to classify partially occluded robustly, models should also be able to localize occluders in images accurately. This improves the explainability of the classification process and enables future research e.g. for parsing scenes with mutually occluding objects. 
Therefore, we propose to test the ability of CompositionalNets and dictionary-based compositional models at occluder localization.
We compute the occlusion score as the log-likelihood ratio between the occluder model and the object model: $\log p(f_p|z^m_p\texttt{=}1)/p(f_p|z^m_p\texttt{=}0)$, where $m=\argmax_m p(F|\theta^m_y)$ is the model that fits the data the best.

\textbf{Quantitative results.} 
We study occluder localization quantitatively on the Occluded-Vehicle dataset using the ground truth segmentation masks of the occluders and the objects. 
Figure \ref{fig:occ-det} shows the ROC curves of CompositionalNets (solid lines) and dictionary-based compositional models (dashed lines) when using the occlusion score to classify each pixel as occluded or non-occluded over all occlusion levels $L1-L3$. 
We evaluate the localization quality only for images that were correctly classified by each model.
The ROC curves show that for both models it is more difficult to localize textured occluders compared to white and noisy occluders. 
Furthermore, it is more difficult to localize natural object occluders compared to textured boxes, likely because of their fine-grained irregular shape.
Overall, CompositionalNets significantly outperform dictionary-based compositional models.
At a false acceptance rate of $0.2$, the performance gain of CompositionalNets is: $12\%$ (white), $19\%$ (noise), $6\%$ (texture) and $8\%$ (objects).

\textbf{Qualitative results.} Figure \ref{fig:occ-quali} qualitatively compares the occluder localization abilities of dictionary-based compositional models and CompositionalNets.
We show images of real and artificial occlusions and the corresponding occlusion scores for all positions $p$ of the feature map $F$. Both models are learned from the \texttt{pool4} feature maps of a VGG-16 network. 
We show more example images in Supplementary D.
Note that we visualize the positive values of the occlusion score after median filtering for illustration purposes (see Supplementary E for unfiltered results).  
We observe that CompositionalNets can localize occluders significantly better compared to the dictionary-based compositional model for real as well as artificial occluders. 
In particular, it seems that dictionary-based compositional models often detect false positive occlusions.
Note how the artificial occluders with white and noise texture are better localized by both models compared to the other occluder types. 

In summary, our qualitative and quantitative occluder localization experiments clearly show that CompositionalNets can localize occluders more accurately compared to dictionary-based compositional models. 
Furthermore, we observe that localizing occluders with variable texture and shape is highly difficult, which could be addressed by developing advanced occlusion models.

	\section{Conclusion}
In this work, we studied the problem of classifying partially occluded objects and localizing occluders in images. 
We found that a standard DCNN does not classify real images of partially occluded objects robustly, even when it has been exposed to severe occlusion during training.
We proposed to resolve this problem by integrating compositional models and DCNNs into a unified model.
In this context, we made the following contributions:

\textbf{Compositional Convolutional Neural Networks.} 
We introduce CompositionalNets, a novel deep architecture with innate robustness to partial occlusion.
In particular we replace the fully connected head in DCNNs with differentiable generative compositional models.

\textbf{Robustness to partial occlusion.} We demonstrate that CompositionalNets can classify partially occluded objects more robustly compared to a standard DCNN and other related approaches, while retaining high discriminative performance for non-occluded images. Furthermore, we show that CompositionalNets can also localize occluders in images accurately, despite being trained with class labels only.

\textbf{Acknowledgement.} This work was partially supported by the Swiss National Science Foundation (P2BSP2.181713) and the Office of Naval Research (N00014-18-1-2119).
	{\small
	\bibliographystyle{plain}
	\bibliography{07_bib}

\begin{thebibliography}{10}

\bibitem{banerjee2005clustering}
Arindam Banerjee, Inderjit~S Dhillon, Joydeep Ghosh, and Suvrit Sra.
\newblock Clustering on the unit hypersphere using von mises-fisher
  distributions.
\newblock {\em Journal of Machine Learning Research}, 6(Sep):1345--1382, 2005.

\bibitem{bienenstock1998compositionality}
Elie Bienenstock and Stuart Geman.
\newblock Compositionality in neural systems.
\newblock In {\em The handbook of brain theory and neural networks}, pages
  223--226. 1998.

\bibitem{bienenstock1997compositionality}
Elie Bienenstock, Stuart Geman, and Daniel Potter.
\newblock Compositionality, mdl priors, and object recognition.
\newblock In {\em Advances in neural information processing systems}, pages
  838--844, 1997.

\bibitem{dai2014unsupervised}
Jifeng Dai, Yi~Hong, Wenze Hu, Song-Chun Zhu, and Ying Nian~Wu.
\newblock Unsupervised learning of dictionaries of hierarchical compositional
  models.
\newblock In {\em Proceedings of the IEEE Conference on Computer Vision and
  Pattern Recognition}, pages 2505--2512, 2014.

\bibitem{deng2009imagenet}
Jia Deng, Wei Dong, Richard Socher, Li-Jia Li, Kai Li, and Li~Fei-Fei.
\newblock Imagenet: A large-scale hierarchical image database.
\newblock In {\em 2009 IEEE conference on computer vision and pattern
  recognition}, pages 248--255. Ieee, 2009.

\bibitem{devries2017improved}
Terrance DeVries and Graham~W Taylor.
\newblock Improved regularization of convolutional neural networks with cutout.
\newblock {\em arXiv preprint arXiv:1708.04552}, 2017.

\bibitem{fawzi2016measuring}
Alhussein Fawzi and Pascal Frossard.
\newblock Measuring the effect of nuisance variables on classifiers.
\newblock Technical report, 2016.

\bibitem{fidler2014}
Sanja Fidler, Marko Boben, and Ales Leonardis.
\newblock Learning a hierarchical compositional shape vocabulary for
  multi-class object representation.
\newblock {\em arXiv preprint arXiv:1408.5516}, 2014.

\bibitem{fodor1988connectionism}
Jerry~A Fodor, Zenon~W Pylyshyn, et~al.
\newblock Connectionism and cognitive architecture: A critical analysis.
\newblock {\em Cognition}, 28(1-2):3--71, 1988.

\bibitem{george2017generative}
Dileep George, Wolfgang Lehrach, Ken Kansky, Miguel L{\'a}zaro-Gredilla,
  Christopher Laan, Bhaskara Marthi, Xinghua Lou, Zhaoshi Meng, Yi~Liu, Huayan
  Wang, et~al.
\newblock A generative vision model that trains with high data efficiency and
  breaks text-based captchas.
\newblock {\em Science}, 358(6368):eaag2612, 2017.

\bibitem{he2016deep}
Kaiming He, Xiangyu Zhang, Shaoqing Ren, and Jian Sun.
\newblock Deep residual learning for image recognition.
\newblock In {\em Proceedings of the IEEE conference on computer vision and
  pattern recognition}, pages 770--778, 2016.

\bibitem{jin2006context}
Ya~Jin and Stuart Geman.
\newblock Context and hierarchy in a probabilistic image model.
\newblock In {\em 2006 IEEE Computer Society Conference on Computer Vision and
  Pattern Recognition (CVPR'06)}, volume~2, pages 2145--2152. IEEE, 2006.

\bibitem{kortylewski2017model}
Adam Kortylewski.
\newblock {\em Model-based image analysis for forensic shoe print recognition}.
\newblock PhD thesis, University\_of\_Basel, 2017.

\bibitem{kortylewski2019compositional}
Adam Kortylewski, Qing Liu, Huiyu Wang, Zhishuai Zhang, and Alan Yuille.
\newblock Combining compositional models and deep networks for robust object
  classification under occlusion.
\newblock {\em arXiv preprint arXiv:1905.11826}, 2019.

\bibitem{kortylewski2016probabilistic}
Adam Kortylewski and Thomas Vetter.
\newblock Probabilistic compositional active basis models for robust pattern
  recognition.
\newblock In {\em British Machine Vision Conference}, 2016.

\bibitem{kortylewski2017greedy}
Adam Kortylewski, Aleksander Wieczorek, Mario Wieser, Clemens Blumer, Sonali
  Parbhoo, Andreas Morel-Forster, Volker Roth, and Thomas Vetter.
\newblock Greedy structure learning of hierarchical compositional models.
\newblock {\em arXiv preprint arXiv:1701.06171}, 2017.

\bibitem{krizhevsky2012imagenet}
Alex Krizhevsky, Ilya Sutskever, and Geoffrey~E Hinton.
\newblock Imagenet classification with deep convolutional neural networks.
\newblock In {\em Advances in neural information processing systems}, pages
  1097--1105, 2012.

\bibitem{li2019aognets}
Xilai Li, Xi~Song, and Tianfu Wu.
\newblock Aognets: Compositional grammatical architectures for deep learning.
\newblock In {\em Proceedings of the IEEE Conference on Computer Vision and
  Pattern Recognition}, pages 6220--6230, 2019.

\bibitem{liao2016learning}
Renjie Liao, Alex Schwing, Richard Zemel, and Raquel Urtasun.
\newblock Learning deep parsimonious representations.
\newblock In {\em Advances in Neural Information Processing Systems}, pages
  5076--5084, 2016.

\bibitem{lin2014microsoft}
Tsung-Yi Lin, Michael Maire, Serge Belongie, James Hays, Pietro Perona, Deva
  Ramanan, Piotr Doll{\'a}r, and C~Lawrence Zitnick.
\newblock Microsoft coco: Common objects in context.
\newblock In {\em European conference on computer vision}, pages 740--755.
  Springer, 2014.

\bibitem{sasikumar2018first}
Dennis Sasikumar, Erik Emeric, Veit Stuphorn, and Charles~E Connor.
\newblock First-pass processing of value cues in the ventral visual pathway.
\newblock {\em Current Biology}, 28(4):538--548, 2018.

\bibitem{simonyan2014very}
Karen Simonyan and Andrew Zisserman.
\newblock Very deep convolutional networks for large-scale image recognition.
\newblock {\em arXiv preprint arXiv:1409.1556}, 2014.

\bibitem{stone2017teaching}
Austin Stone, Huayan Wang, Michael Stark, Yi~Liu, D~Scott~Phoenix, and Dileep
  George.
\newblock Teaching compositionality to cnns.
\newblock In {\em Proceedings of the IEEE Conference on Computer Vision and
  Pattern Recognition}, pages 5058--5067, 2017.

\bibitem{tabernik2016towards}
Domen Tabernik, Matej Kristan, Jeremy~L Wyatt, and Ale{\v{s}} Leonardis.
\newblock Towards deep compositional networks.
\newblock In {\em 2016 23rd International Conference on Pattern Recognition
  (ICPR)}, pages 3470--3475. IEEE, 2016.

\bibitem{tang2018deeply}
Wei Tang, Pei Yu, and Ying Wu.
\newblock Deeply learned compositional models for human pose estimation.
\newblock In {\em Proceedings of the European Conference on Computer Vision
  (ECCV)}, pages 190--206, 2018.

\bibitem{tang2017towards}
Wei Tang, Pei Yu, Jiahuan Zhou, and Ying Wu.
\newblock Towards a unified compositional model for visual pattern modeling.
\newblock In {\em Proceedings of the IEEE International Conference on Computer
  Vision}, pages 2784--2793, 2017.

\bibitem{vaziri2014channel}
Siavash Vaziri, Eric~T Carlson, Zhihong Wang, and Charles~E Connor.
\newblock A channel for 3d environmental shape in anterior inferotemporal
  cortex.
\newblock {\em Neuron}, 84(1):55--62, 2014.

\bibitem{von1987synaptic}
Ch~von~der Malsburg.
\newblock Synaptic plasticity as basis of brain organization.
\newblock {\em The neural and molecular bases of learning}, 411:432, 1987.

\bibitem{wang2017detecting}
Jianyu Wang, Cihang Xie, Zhishuai Zhang, Jun Zhu, Lingxi Xie, and Alan Yuille.
\newblock Detecting semantic parts on partially occluded objects.
\newblock {\em British Machine Vision Conference}, 2017.

\bibitem{wang2015unsupervised}
Jianyu Wang, Zhishuai Zhang, Cihang Xie, Vittal Premachandran, and Alan Yuille.
\newblock Unsupervised learning of object semantic parts from internal states
  of cnns by population encoding.
\newblock {\em arXiv preprint arXiv:1511.06855}, 2015.

\bibitem{wang2017visual}
Jianyu Wang, Zhishuai Zhang, Cihang Xie, Yuyin Zhou, Vittal Premachandran, Jun
  Zhu, Lingxi Xie, and Alan Yuille.
\newblock Visual concepts and compositional voting.
\newblock {\em arXiv preprint arXiv:1711.04451}, 2017.

\bibitem{xiang2014beyond}
Yu~Xiang, Roozbeh Mottaghi, and Silvio Savarese.
\newblock Beyond pascal: A benchmark for 3d object detection in the wild.
\newblock In {\em IEEE Winter Conference on Applications of Computer Vision},
  pages 75--82. IEEE, 2014.

\bibitem{xiao2019tdapnet}
Mingqing Xiao, Adam Kortylewski, Ruihai Wu, Siyuan Qiao, Wei Shen, and Alan
  Yuille.
\newblock Tdapnet: Prototype network with recurrent top-down attention for
  robust object classification under partial occlusion.
\newblock {\em arXiv preprint arXiv:1909.03879}, 2019.

\bibitem{yamane2008neural}
Yukako Yamane, Eric~T Carlson, Katherine~C Bowman, Zhihong Wang, and Charles~E
  Connor.
\newblock A neural code for three-dimensional object shape in macaque
  inferotemporal cortex.
\newblock {\em Nature neuroscience}, 11(11):1352, 2008.

\bibitem{yun2019cutmix}
Sangdoo Yun, Dongyoon Han, Seong~Joon Oh, Sanghyuk Chun, Junsuk Choe, and
  Youngjoon Yoo.
\newblock Cutmix: Regularization strategy to train strong classifiers with
  localizable features.
\newblock {\em arXiv preprint arXiv:1905.04899}, 2019.

\bibitem{zhang2018interpretable}
Quanshi Zhang, Ying Nian~Wu, and Song-Chun Zhu.
\newblock Interpretable convolutional neural networks.
\newblock In {\em Proceedings of the IEEE Conference on Computer Vision and
  Pattern Recognition}, pages 8827--8836, 2018.

\bibitem{zhang2018deepvoting}
Zhishuai Zhang, Cihang Xie, Jianyu Wang, Lingxi Xie, and Alan~L Yuille.
\newblock Deepvoting: A robust and explainable deep network for semantic part
  detection under partial occlusion.
\newblock In {\em Proceedings of the IEEE Conference on Computer Vision and
  Pattern Recognition}, pages 1372--1380, 2018.

\bibitem{hongru}
Hongru Zhu, Peng Tang, Jeongho Park, Soojin Park, and Alan Yuille.
\newblock Robustness of object recognition under extreme occlusion in humans
  and computational models.
\newblock {\em CogSci Conference}, 2019.

\bibitem{zhu2008}
Long~Leo Zhu, Chenxi Lin, Haoda Huang, Yuanhao Chen, and Alan Yuille.
\newblock Unsupervised structure learning: Hierarchical recursive composition,
  suspicious coincidence and competitive exclusion.
\newblock In {\em Computer vision--eccv 2008}, pages 759--773. Springer, 2008.

\end{thebibliography}
	}
\end{document}